\documentclass[sn-mathphys-num]{sn-jnl}% Math and Physical Sciences Numbered Reference Style 
\usepackage{graphicx}%
\usepackage{multirow}%
\usepackage{amsmath,amssymb,amsfonts}%
\usepackage{amsthm}%
\usepackage{mathrsfs}%
\usepackage[title]{appendix}%
\usepackage{xcolor}%
\usepackage{textcomp}%
\usepackage{manyfoot}%
\usepackage{booktabs}%
\usepackage{algorithm}%
\usepackage{algorithmicx}%
\usepackage{algpseudocode}%
\usepackage{listings}%
\theoremstyle{thmstyleone}%
\theoremstyle{thmstyletwo}%

\theoremstyle{thmstylethree}%

\begin{document}

\title[Article Title]{Knowledge Elicitation with Large Language Models for Interpretable Cancer Stage Identification from Pathology Reports}


\author[1]{\fnm{Yeawon} \sur{Lee}}\email{yl3427@drexel.edu}

\author*[1]{\fnm{Christopher C.} \sur{Yang}}\email{chris.yang@drexel.edu}

\author[2]{\fnm{Chia-Hsuan} \sur{Chang}}\email{shane.chang.tw@gmail.com}

\author[3]{\fnm{Grace} \sur{Lu-Yao}}\email{Grace.LuYao@jefferson.edu}

\affil[1]{\orgdiv{College of Computing and Informatics}, \orgname{Drexel University}, \orgaddress{\city{Philadelphia}, \postcode{19104}, \state{PA}, \country{Country}}}
\affil[2]{\orgdiv{Yale School of Medicine}, \orgname{Yale University}, \orgaddress{\city{New Haven}, \postcode{06510}, \state{CT}, \country{USA}}}
%\affil*[3]{\orgdiv{College of Computing and Informatics}, \orgname{Drexel University}, \orgaddress{\city{Philadelphia}, \postcode{19104}, \state{PA}, \country{Country}}}
\affil[3]{\orgdiv{Department of Medical Oncology, Sidney Kimmel Cancer Center}, \orgname{Thomas Jefferson University}, \orgaddress{\city{Philadelphia}, \postcode{19107}, \state{PA}, \country{USA}}}

\abstract{Cancer staging is critical for patient prognosis and treatment planning, yet extracting pathologic TNM staging from unstructured pathology reports poses a persistent challenge. Existing natural language processing (NLP) and machine learning (ML) strategies often depend on large annotated datasets, limiting their scalability and adaptability. 
%% <MODIFICATION START>
In this study, we introduce two Knowledge Elicitation methods designed to overcome these limitations by enabling large language models (LLMs) to induce and apply domain-specific rules for cancer staging. The first, Knowledge Elicitation with Long-Term Memory (KEwLTM), uses an iterative prompting strategy to derive staging rules directly from unannotated pathology reports, without requiring ground-truth labels. The second, Knowledge Elicitation with Retrieval-Augmented Generation (KEwRAG), employs a variation of RAG where rules are pre-extracted from relevant guidelines in a single step and then applied, enhancing interpretability and avoiding repeated retrieval overhead. We leverage the ability of LLMs to apply broad knowledge learned during pre-training to new tasks. Using breast cancer pathology reports from the TCGA dataset, we evaluate their performance in identifying T and N stages, comparing them against various baseline approaches on two open-source LLMs. Our results indicate that KEwLTM outperforms KEwRAG when Zero-Shot Chain-of-Thought (ZSCOT) inference is effective, whereas KEwRAG achieves better performance when ZSCOT inference is less effective. Both methods offer transparent, interpretable interfaces by making the induced rules explicit. These findings highlight the promise of our Knowledge Elicitation methods as scalable, high-performing solutions for automated cancer staging with enhanced interpretability, particularly in clinical settings with limited annotated data.}
%% <MODIFICATION END>


\maketitle

\section{Introduction}\label{sec1}
Cancer staging is critical for determining the prognosis, treatment plan, and overall clinical management for cancer patients. The American Joint Committee on Cancer (AJCC) TNM classification system is widely used to stage cancer based on three key factors: Tumor size (T), regional lymph Node involvement (N), and the presence or absence of distant Metastasis (M). Obtaining pathologic TNM staging information usually requires manually parsing and extracting the relevant details from pathology reports, which are typically provided in free-text format. This unstructured format makes it difficult for healthcare providers to quickly access stage information, creating a need for natural language processing (NLP) solutions.

In recent years, NLP techniques~\cite{Odisho2020-vy,De_Angeli2021-ro,Gao2018-ex,Gao2019-qc,Wu2020-du} have been used to automate the extraction of cancer stage information from free-text pathology reports. However, traditional NLP, machine learning (ML), and deep learning methods typically require large, annotated datasets for training, which can be expensive and time-consuming to create. Furthermore, these methods often struggle to perform well when applied to reports from different hospitals or institutions due to variations in reporting styles and terminology without retraining or adaptation. Therefore, there is a critical need for a scalable solution for cancer stage classification that does not rely heavily on labor-intensive labeled datasets, especially in clinical settings.

%% revision start (0512)
The emergence of large language models (LLMs) such as Llama, Mixtral, and GPT has introduced a new era in NLP. LLMs have demonstrated the ability to apply their knowledge across diverse tasks, often without requiring explicit training on each new dataset. This capability stems from their extensive pre-training on vast corpora. During this pre-training, LLMs can learn general medical knowledge, including the fundamental rules of cancer staging, from publicly available resources like medical textbooks and guidelines~\cite{Vaswani2017-gk,Carlini2023-ys}. Previous studies~\cite{Chang2024-qd,Chang2024-qu} have shown the effectiveness of using LLMs for cancer stage classification tasks through prompting. However, prompting approaches that rely solely on the LLM's existing general knowledge may achieve sub-optimal performance when addressing context-specific clinical details. Actual patient pathology reports, which contain such details, are typically part of the electronic health record and protected under privacy regulations. Consequently, LLMs cannot access them during pre-training. This lack of exposure to real-world clinical narratives, with their inherent variability and specialized terminology, can limit the LLM's ability to accurately apply its general knowledge in these specific contexts.
%% revision end (0512)

%% <MODIFICATION START>
To address these limitations, while building upon established techniques like iterative prompting and retrieval-augmented generation (RAG), we propose two Knowledge Elicitation approaches with specific adaptations for clinical utility: Knowledge Elicitation with Long-Term Memory (KEwLTM) and Knowledge Elicitation with Retrieval-Augmented Generation (KEwRAG).

KEwLTM enables LLMs to derive domain-specific knowledge directly from a small number of unannotated pathology reports. A key aspect of KEwLTM is its label-free induction process. It does not rely on ground-truth labels or human annotations. Instead, the model iteratively induces and refines high-level staging rules from the content of the instance reports themselves, storing these rules in a persistent long-term memory. This approach is particularly valuable in clinical contexts where large annotated datasets are scarce or restricted. The explicit rules also enhance interpretability.

In contrast, KEwRAG adapts the standard RAG framework by front-loading the rule extraction. Instead of retrieving and appending raw text chunks to the LLM for each query, KEwRAG first retrieves relevant information from an external source (e.g., clinical guidelines) once. It then prompts the LLM to synthesize these retrieved texts into a concise, structured set of rules. This stable set of rules is then used for subsequent inferences, eliminating repeated retrieval overhead and providing a more coherent, auditable knowledge base that clinicians can easily review and validate.

Both methods are designed for practical integration in limited-data clinical settings, aiming to be usable locally and generate domain-specific knowledge in a compact, reusable, and interpretable form. Our findings suggest these adapted methods can improve cancer stage identification accuracy while providing clinicians with transparent insights into how predictions are made, thereby promoting effective human-AI collaboration in healthcare.
%% <MODIFICATION END>

\section{Related Work}\label{sec2}
In recent years, natural language processing (NLP) and machine learning (ML) have been instrumental in developing automated systems for extracting cancer stage information from free-text pathology reports. Odisho et al.~\cite{Odisho2020-vy} utilized contextual token embeddings with logistic regression, AdaBoost, and random forests for pathologic stage classification. Angeli et al.~\cite{De_Angeli2021-ro} applied active learning to select training samples and then trained a convolutional neural network for stage identification. Gao et al.~\cite{Gao2018-ex, Gao2019-qc} developed a hierarchical network to learn representations from words to complete reports, demonstrating its effectiveness in tumor grade classification using SEER data. Wu et al.~\cite{Wu2020-du} leveraged attention-based graph convolution networks, using multi-source knowledge graphs for improved TNM stage identification on TCGA data.

However, the adaptability of these ML and deep learning models remains limited, particularly because they rely on labeled datasets, which are often expensive to produce and constrained in scope. This limitation makes it difficult for these models to adapt to data that significantly differ from their training datasets, such as the diverse formats used in pathology reports across different medical facilities and the varying rules for different cancer types.

Foundation models, such as pre-trained language models, are designed to be more adaptable by training on vast amounts of unstructured data~\cite{Vaswani2017-gk}. This large-scale, raw corpus enables them to learn from diverse, real-world scenarios without the need for labor-intensive labeling. Their broad exposure allows them to encode a wide array of knowledge across their vast parameter space~\cite{Carlini2023-ys}, equipping them to handle novel tasks and complex real-world data more readily, making them powerful tools for various healthcare applications.

Kefeli et al.~\cite{Kefeli2023-ec} leveraged the power of pre-trained language models by fine-tuning a clinical-specific model, Clinical-BigBird~\cite{Li2022-lw}, for TNM classification using reports from the TCGA project. Their fine-tuned model achieved strong performance both on the TCGA test reports and on an independent set of real-world pathology reports, collected from Columbia University Irving
Medical Center. Furthermore, Chang et al.~\cite{Chang2024-qd} investigated open-source clinical large language model (Med42-70B~\cite{Christophe2024-wz}) in cancer stage identification. They found that adopting the zero-shot chain-of-thought (ZSCOT) prompting approach, without requiring any training samples, reached comparable performance in TNM classification task to fine-tuned BERT-based models. Recently, Chang et al.~\cite{Chang2024-qu} introduced a new iterative prompting workflow called ensemble reasoning (EnsReas) which aims to enhance predictive performance and consistency by extending the ZSCOT approach. While promising for improving performance and consistency, EnsReas requires a high volume of API calls to the LLM, making it inefficient and costly. 

%% <MODIFICATION START>
Our current study explores two efficient prompting approaches that, while building on concepts like iterative prompting and RAG, introduce specific adaptations to induce domain-specific knowledge and incorporate it into the LLM’s generated reasoning. These adaptations aim to enhance predictive performance and interpretability in ways that differ from standard applications. Specifically, KEwLTM derives knowledge through a label-free iterative rule induction process directly from a small number of unannotated pathology reports, storing it in long-term memory. This contrasts with many iterative prompting methods that may still require some form of labeled data or explicit external feedback. KEwRAG, on the other hand, modifies the typical RAG process by retrieving relevant information once at the outset and then prompting the LLM to synthesize this into an explicit, interpretable set of rules, which are then applied throughout the inference process. This differs from standard RAG that often appends raw retrieved text per query. Both methods prioritize the generation of an explicit, reviewable rule set.
%% <MODIFICATION END>

\section{Methods \& Materials}\label{sec3}
    \subsection{Dataset}\label{subsec3-1}
    We utilize a dataset of free-text pathology reports from the Cancer Genomic Atlas (TCGA) project of the National Cancer Institute (NCI) which have been pre-processed as described in~\cite{Kefeli2024-gk}. The metadata associated with each report can be found on the NCI Genomic Data Commons (GDC) portal, and each report is associated with one or more of components of the pathology stage (i.e., T, N, or M status). Previous studies~\cite{Chang2024-qd,Chang2024-qu,Kefeli2023-ec} have used this dataset to investigate various NLP and LLM methods.

    In this study, we focus on a subset of pathology reports for patients with breast cancer (BRCA). Globally, breast cancer remains one of the most frequently diagnosed malignancies, underscoring the importance of developing accurate and interpretable staging solutions for this disease. Furthermore, BRCA is particularly well-represented in our dataset, comprising around 800 reports, compared to the second-largest subset—lung cancer—with only 477. This larger sample size not only reflects the clinical significance of breast cancer but also provides a more robust basis for developing and evaluating our proposed approaches. Also, we only evaluate the performance for T and N stage identification because the M stage for most reports is undetermined (i.e., Mx) or not available. Table~\ref{tab: dataset statistics} provides the class distribution of T and N category.

    \setlength{\tabcolsep}{15pt} 
    \begin{table}[ht]
    \centering
    \caption{Distribution of T and N category of BRCA pathology reports}
    \label{tab: dataset statistics}
    \begin{tabular}{l c c c c c}  % 6 columns total
    \toprule
    T Category & T1 & T2 & T3 & T4 & Total \\
    \cmidrule(lr){2-6}
              & 468 & 188 & 108 & 36 & 800 \\
    \midrule
    N Category & N0 & N1 & N2 & N3 & Total \\
    \cmidrule(lr){2-6}
              & 316 & 300 & 110 & 74 & 800 \\
    \bottomrule
    \end{tabular}
    \end{table}

    \subsection{Knowledge Elicitation with Long-Term Memory (KEwLTM)}\label{subsec3-2}

    \begin{figure}[h]
    \centering
    \includegraphics[width=0.9\textwidth]{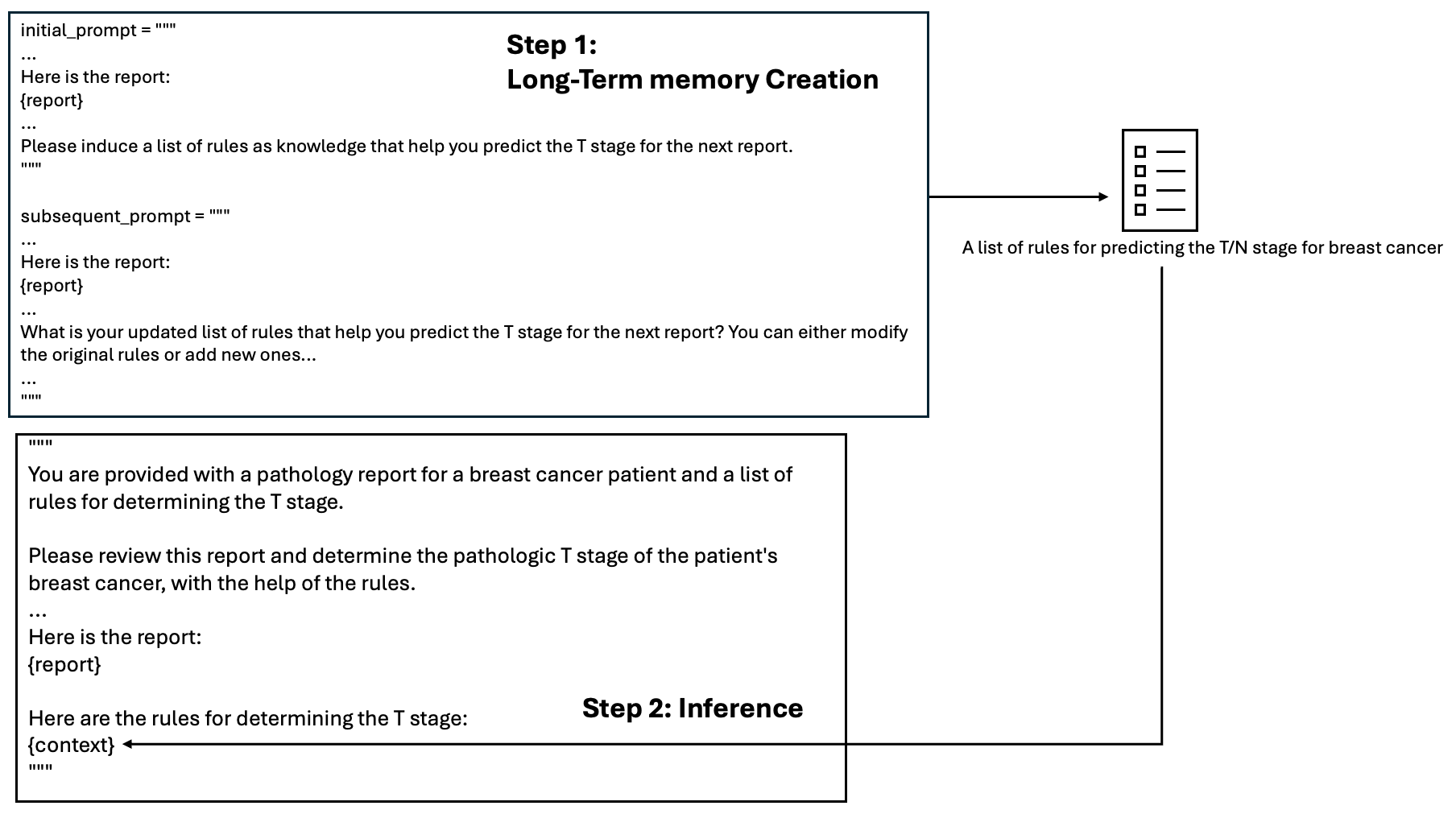}
    \caption{Overview of the KEwLTM workflow}\label{fig: KEwLTM illustration}
    \end{figure}

    %% revision start (0512)
    In the context of LLMs, ``memorization'' refers to the knowledge embedded within the LLM's parameters during pre-training~\cite{Carlini2023-ys}. Through this pre-training on vast public datasets, LLMs acquire general knowledge of cancer staging principles (e.g., the relationship between tumor size and T-stage, or the number of positive lymph nodes and N-stage) from publicly available medical literature, textbooks, and guidelines, such as those from the AJCC. Evidence for this baseline knowledge is apparent when LLMs, if prompted appropriately, can generate staging rules that align with established clinical criteria. However, this general understanding does not translate to expertise in interpreting the specific content and diverse, often nuanced, phrasing found in real-world pathology reports. These reports contain protected health information and are typically not part of the large-scale corpora used for pre-training. This lack of direct exposure means that while an LLM might know the general logic of staging, it is unfamiliar with applying it to the heterogeneous nature of actual patient notes.
    
    To bridge this gap between general pre-trained knowledge and the specifics of pathology reports, we propose Knowledge Elicitation with Long-Term Memory (KEwLTM). As illustrated in Figure~\ref{fig: KEwLTM illustration}, KEwLTM guides the LLM to induce domain-specific rules directly from a small number of unlabelled pathology reports. These induced rules are then stored and refined in a persistent long-term memory. Algorithm~\ref{alg:KEwLTM} describes this workflow. We define the LLM as an autoregressive model $\mathcal{A}$ that generates text based on inputs, instructions, and provided contexts. The core idea of KEwLTM is to leverage the LLM's foundational, pre-trained understanding of cancer staging and augment it with contextually relevant rules elicited from sample reports. This induced knowledge, stored as long-term memory $\mathcal{M}$, then provides the LLM with enhanced, specialized context for subsequent cancer stage identification tasks. Our goal is to enable the LLM to apply its general knowledge to domain-specific data through $\mathcal{M}$ as its elicited memory.
    %% revision end (0512)
    
    To achieve this, we split the dataset into a training set $D_{\text{train}}$ (100 reports) and a test set $D_{\text{test}}$ (700 reports). Initially, when no long-term memory exists, the LLM $\mathcal{A}$ generates $\mathcal{M}$ using the prompt template $T_{\text{elicit}}$, which presents the first instance report in $D_{\text{train}}$. $T_{\text{elicit}}$ directs LLM to generate reasoning ($r$), predict the cancer stage ($\hat{y}$), and extract a list of rules ($m$) that serve as the initial long-term memory.
    
    With the initialized $\mathcal{M}$, subsequent reports in $D_{\text{train}}$ are processed using another prompt template $T_{\text{update}}$. This template, $T_{\text{update}}$, takes the current long-term memory $\mathcal{M}$ and the next training report $x^{(\text{train})}$ as inputs to formulate a prompt. This prompt directs the LLM to generate reasoning ($r$), predict the cancer stage ($\hat{y}$), and propose an updated list of rules ($m$) for the long-term memory, potentially by adding to, modifying, or deleting existing rules. The long-term memory ($\mathcal{M}$) is updated iteratively until all 100 training reports are consumed. Due to the inherent randomness of LLM generation, updates to $\mathcal{M}$ are accepted only if the newly generated rules ($m$) closely match the existing long-term memory ($\mathcal{M}$). This similarity is measured using edit distance (Levenshtein distance), ensuring that $\mathcal{M}$ evolves in a controlled and incremental manner.

    %% <MODIFICATION START>
    It is important to note that the term "training" here refers to the process of long-term memory induction (for $\mathcal{M}$) from unannotated reports and differs from its traditional meaning in machine learning. In this method, no labels, references, or ground truth are required during this induction phase (although reference labels are available in the dataset and used for subsequent evaluation on the test set). Instead, a portion of the reports is provided to the model, and it elicits $\mathcal{M}$ while also making stage predictions. These predictions during this long-term memory induction phase are considered auxiliary outputs and are not used for the final performance evaluation, as the focus of this phase is on eliciting $\mathcal{M}$.
    %% <MODIFICATION END>
    
    Once $\mathcal{M}$ is finalized, we proceed to the test phase. The prompt template $T_{\text{inference}}$ takes a test report $x^{(\text{test})}$ and the finalized long-term memory $\mathcal{M}$ as inputs to formulate a prompt. This prompt directs the LLM to generate reasoning ($r$) and predict the cancer stage ($\hat{y}$) for the report.

    \begin{algorithm}
        \caption{KEwLTM}
        \label{alg:KEwLTM}
        \begin{algorithmic}[1]
            \Require LLM $\mathcal{A}$, Training set $D_{\text{train}}$, Test set $D_{\text{test}}$, Threshold $\delta$, Prompt templates $T_{\text{elicit}}$, $T_{\text{update}}$, and $T_{\text{inference}}$
            \State \textbf{Step 1: Long-Term Memory Creation}
            \State Initialize long-term memory $\mathcal{M} \gets \emptyset$
            \For{each $x^{(\text{train})} \in D_{\text{train}}$}
                \If{$\mathcal{M} == \emptyset$}
                    \State $r, \hat{y}, m \gets \mathcal{A}(T_{\text{elicit}}(x^{(\text{train})}))$
                    \State $\mathcal{M} \gets m$
                \Else
                    \State $r, \hat{y}, m \gets \mathcal{A}(T_{\text{update}}(x^{(\text{train})}, \mathcal{M}))$
                    \State $d \gets \text{Edit\_Distance}(m, \mathcal{M})$
                    \If{$d \leq \delta$}
                        \State $\mathcal{M} \gets m$
                    \EndIf
                \EndIf
            \EndFor
            \State \Return $\mathcal{M}$
            \State
            \State \textbf{Step 2: Inference}
            \For{each $x^{(\text{test})} \in D_{\text{test}}$}
                \State $r, \hat{y} \gets \mathcal{A}(T_{\text{inference}}(x^{(\text{test})}, \mathcal{M}))$
            \EndFor
        \end{algorithmic}
    \end{algorithm}
    
    \subsection{Knowledge Elicitation with Retrieval-Augmented Generation}\label{subsec3-3}

    \begin{figure}[h]
    \centering
    \includegraphics[width=0.9\textwidth]{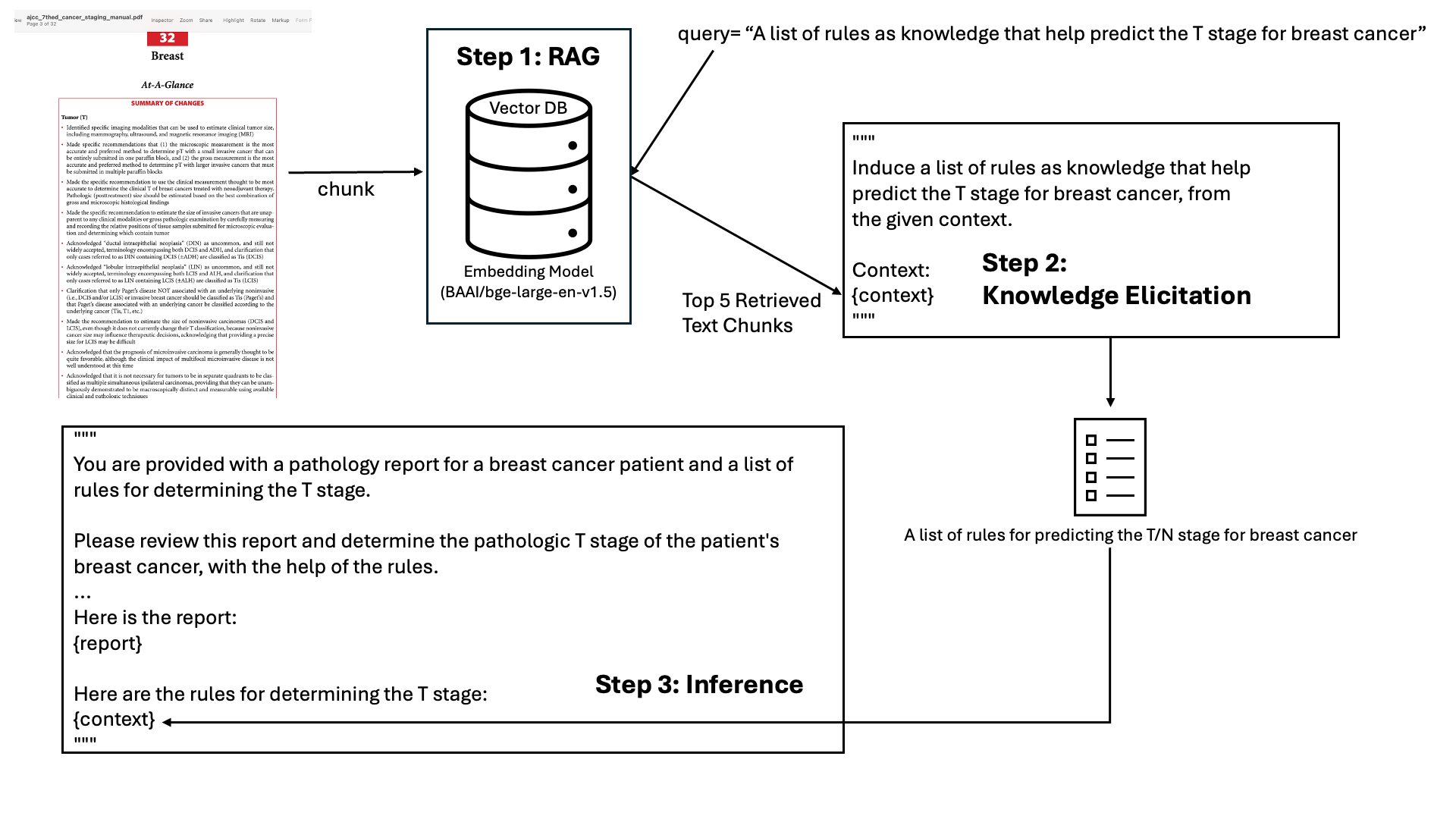}
    \caption{Overview of the KEwRAG workflow, where relevant text chunks are retrieved before rules are elicited.}\label{fig: KEwRAG illustration}
    \end{figure}

     %% <MODIFICATION START>
    In this approach, the cancer staging rules, $\mathcal{K}$, are derived from external domain resources rather than training reports. Since our TCGA dataset’s annotations align with the 7th edition of the AJCC cancer staging manual, that edition serves as the knowledge source. As illustrated in Figure~\ref{fig: KEwRAG illustration}, KEwRAG first retrieves relevant content from the AJCC manual. Then, instead of directly using these retrieved raw text chunks for each inference query, KEwRAG prompts the LLM to synthesize these chunks into a structured, explicit set of staging rules $\mathcal{K}$. This rule induction occurs once. This set $\mathcal{K}$ is then used as a stable knowledge context for all subsequent inferences, which offers advantages in terms of interpretability, auditability, and reduced computational overhead from repeated retrievals.
    %% <MODIFICATION END>

    %% Revision start (0512)
    Algorithm~\ref{alg:KEwRAG} illustrates the KEwRAG workflow in three main steps. First, a retrieval module $\mathcal{R}$ selects the top $k$ most relevant text chunks (e.g., short paragraphs or sections containing staging definitions) from the knowledge source based on a query $q$. We set $k=5$ for our experiments as a heuristic choice, aiming to provide sufficient context from the guidelines without excessive input length or computational overhead; this parameter could be tuned in future work. We employed a relatively general query formulation, such as the example shown in Figure~\ref{fig: KEwRAG illustration} ('A list of rules as knowledge that help predict the T stage for breast cancer'), adapted for T and N stages respectively. We acknowledge that we did not systematically experiment with different query structures or values of $k$ in this study, and further optimization of retrieval parameters like $k$ and query formulation could be explored in future work.

     The retrieved chunks $\mathcal{C}$ are concatenated and provided as context to the LLM $\mathcal{A}$. The prompt template $T_{\text{elicit}}$ takes the retrieved text chunks $\mathcal{C}$ as input to formulate a prompt. This prompt directs the LLM to synthesize these chunks into a structured set of staging rules, denoted as knowledge $\mathcal{K}$. Unlike KEwLTM, which iteratively refines rules by learning from multiple example reports, KEwRAG was designed to perform this rule synthesis in a single pass. This was a design choice primarily for simplicity and efficiency within the retrieval-based framework, aiming to accurately distill the information present in the retrieved expert-authored text chunks in one step. Finally, during inference, the prompt template $T_{\text{inference}}$ takes an input report $x$ and the elicited knowledge $\mathcal{K}$ as inputs to formulate a prompt. This prompt directs the LLM to generate reasoning ($r$) and predict the cancer stage ($\hat{y}$) for the report.
     %% Revision end (0512)
    
    \begin{algorithm}
        \caption{KEwRAG}
        \label{alg:KEwRAG}
        \begin{algorithmic}[1]
            \Require LLM $\mathcal{A}$, Full dataset $D$, RAG module $\mathcal{R}$, RAG query $q$, Prompt templates $T_{\text{elicit}}$ and $T_{\text{inference}}$
            \State \textbf{Step 1: RAG}
            \State Retrieve text chunks $\mathcal{C} \gets \mathcal{R}(q, \text{AJCC Guideline})$
            \State \textbf{Step 2: Knowledge Elicitation}
            \State Elicit knowledge $\mathcal{K} \sim \mathcal{A}(T_{\text{elicit}}(\mathcal{C}))$
            \State \textbf{Step 3: Inference}
            \For{each $x \in D$}
                \State $r, \hat{y} \sim \mathcal{A}(T_{\text{inference}}(x, \mathcal{K}))$
            \EndFor
        \end{algorithmic}
    \end{algorithm}
        
    \subsection{Experimental Settings}\label{subsec3-4}

        \subsubsection{Model Setup}\label{subsubsec3-4-1} 
        
        Our selection of primary large language models (LLMs) for this study was guided by several key considerations. A fundamental requirement was the use of open-source models that could be deployed locally, ensuring data privacy and security when processing sensitive clinical text. Among suitable open-source options, we chose Mixtral-8x7B-Instruct-v0.1\footnote{Available at: \url{https://huggingface.co/mistralai/Mixtral-8x7B-Instruct-v0.1}}~\cite{Jiang2024-ev} and Llama3-Med42-70B\footnote{Available at: \url{https://huggingface.co/m42-health/Llama3-Med42-70B}}~\cite{Christophe2024-wz}. While other models, including larger proprietary systems (e.g., GPT-4 class) and earlier open-source Llama versions, were considered, they were not selected primarily due to restrictions on local deployment, higher computational or financial resource demands beyond our available infrastructure, or less optimal context window sizes for our specific application of processing detailed pathology reports. We served both selected models on four NVIDIA A40 GPUs.
        
        Mixtral-8x7B-Instruct-v0.1~\cite{Jiang2024-ev} is a pre-trained generative model utilizing a sparse mixture-of-experts architecture. Its large 32K-token context window was a key advantage for our work, offering the capacity to process lengthy pathology reports effectively. By contrast, Llama3-Med42-70B~\cite{Christophe2024-wz}, developed by M42 Health, is an open-access clinical LLM built upon the Llama3 architecture. It has been instruction-tuned on extensive medical data, which is intended to enhance its performance on healthcare-related language tasks. This model supports an 8K-token context window.
        
        For our RAG pipeline, we loaded NV-Embed-v2\footnote{Available at: \url{https://huggingface.co/nvidia/NV-Embed-v2}}~\cite{lee2025nvembedimprovedtechniquestraining} using half precision (FP16). This format stores the model's parameters using 16-bit floating-point numbers (instead of the standard 32-bit), which reduces GPU memory usage and can accelerate inference speed on compatible hardware, making deployment more efficient.
        
        It is important to acknowledge that pre-trained LLMs, including those employed in this study, can inherit and potentially amplify biases present in their vast training corpora. Such biases might relate to demographic factors, geographic origins, institutional reporting practices, or underrepresentation of certain patient populations. The propagation of these biases could lead to disparities or inaccuracies if the models are applied broadly in diverse clinical settings without careful validation and mitigation. While specific mitigation strategies are beyond the scope of this initial methodological investigation, we recognize their critical importance for any future real-world deployment.

        \subsubsection{Baselines}\label{subsubsec3-4-2}
        We compare KEwLTM and KEwRAG against two additional baselines: Zero-Shot Chain-of-Thought (ZSCOT) and Retrieval-Augmented Generation (RAG). ZSCOT~\cite{Kojima2024-na} is a method where a language model is prompted to explain its reasoning steps without prior examples. We include it as a baseline because it is widely used, straightforward, and offers interpretability by having the model articulate its reasoning—a key requirement in healthcare applications.

        RAG, on the other hand, retrieves relevant chunks of information from external sources and uses them as context to generate answers. This not only reduces hallucinations but also increases transparency: users can inspect the retrieved content, which makes the reasoning more interpretable. Moreover, since one of our methods (KEwRAG) adapts the RAG approach by pre-extracting rules from retrieved content, including standard RAG (which typically appends raw chunks per query) as a baseline allows us to directly compare our proposed method with the conventional retrieval-augmented framework and highlight the benefits of our rule synthesis step.

        \subsubsection{Performance Metrics}\label{subsubsec3-4-3}
       
        % revision start (0512)
        We frame the cancer stage identification task as two separate multi-class classification problems: one for the T category (classes: T1, T2, T3, T4) and one for the N category (classes: N0, N1, N2, N3). The reference labels for these tasks are provided in the TCGA dataset as document-level annotations for each pathology report (e.g., a single 'T1' label and a single 'N0' label per report). 
        
        Although LLMs typically generate free-form text, for robust evaluation, we constrained their final output to a structured JSON format. This was achieved by serving the LLMs via the vLLM package~\cite{kwon2023efficientmemorymanagementlarge} using the 'lm-format-enforcer' backend\footnote{lm-format-enforcer available at \url{https://github.com/noamgat/lm-format-enforcer}}. We defined a JSON schema that required the LLM to produce two fields: 
        1) a 'reasoning' field, containing the textual step-by-step explanation for its prediction; and 
        2) a 'stage' field, which was strictly constrained to output one of the predefined class labels (e.g., 'T1', 'T2', 'T3', or 'T4' for T-category tasks, and similarly for N-category tasks).
        
        This structured output ensures that we obtain an unambiguous, predicted class label from the LLM for each report. While the `reasoning` string is preserved for interpretability and qualitative error analysis (as discussed in Section~\ref{subsec4-2}), our quantitative performance evaluation focuses solely on automatically comparing the value in the `stage` field of the LLM's JSON output against the corresponding document-level reference label from the TCGA dataset. This allows for direct and consistent calculation of the performance metrics.

        To evaluate performance, we use the metrics of precision, recall, and F1 score, defined as:
        \begin{equation}
            precision = \frac{TP}{TP+FP},
        \end{equation}
        \begin{equation}
            recall = \frac{TP}{TP+FN},
        \end{equation}
        \begin{equation}
            F1 = 2 \times \frac{precision \times recall}{precision + recall},
        \end{equation}
        \noindent where TP, FP, and FN represent the number of true positives, false positives, and false negatives, respectively.
        
        Given the imbalanced class distribution within both T and N categories (as shown in Table~\ref{tab: dataset statistics}), simply reporting overall accuracy could be misleading. The performance on less frequent classes (e.g., T4, N3) is clinically important and should not be overshadowed by performance on more frequent classes. Therefore, we calculate precision, recall, and F1 score for each individual class and then compute the macro-average for each metric. Macro-averaging calculates the metric independently for each class and then takes the unweighted mean, thus treating all classes as equally important regardless of their frequency. We report these macro-averaged scores for comparing the different prompting approaches.
        
        Note that for the KEwLTM method specifically, the evaluation process involves splitting the dataset. To ensure robust results, we created eight distinct random splits, each comprising 100 reports for the memory induction phase ('training') and the remaining 700 reports for the evaluation phase ('testing'). We evaluated KEwLTM on each of the eight test splits and report the average macro-precision, macro-recall, and macro-F1 scores across these eight runs. The results for the other methods (ZSCOT, RAG, KEwRAG) are based on evaluating all 800 reports once, as they do not require a separate training/induction split in the same manner.
        
        Although we initially set aside 100 reports for training, we did not always use them all. We tracked how the memory length changed at different training-report counts (up to 100) under two conditions: with no edit distance threshold, and with an edit distance threshold. When we set the threshold to 80 (here, the score indicates “similarity,” meaning the edit distance must be below 20), the memory grew more slowly but stabilized at around 40 training reports for both T and N categories. 

        Therefore, we used 40 training reports to build KEwLTM’s memory in our final evaluations. A detailed sensitivity analysis regarding the number of training reports and the rationale for the edit distance threshold, including graphical illustrations of their impact on performance and memory evolution, is provided in Appendix~\ref{app:sensitivity}. Note that this “optimal” threshold, as well as the number of training reports used, can still vary depending on the order in which the training reports are introduced, even when using the same set of 100 reports.
        
\section{Experimental Results}\label{sec4}

\begin{table*}[ht]
\centering
\caption{Comparison of T and N stage classification performance.
For \textbf{KEwLTM} we report the mean over eight runs $\pm$ standard deviation.
The right‑most column is the macro‑averaged F$_1$ across both T and N categories.}
\label{tab:performance-comparison}
\resizebox{\textwidth}{!}{%
\begin{tabular}{@{}llccccccc@{}}
\toprule
 & & \multicolumn{3}{c}{T category} &
   \multicolumn{3}{c}{N category} &
   \multirow{2}{*}{Macro F$_1$ (T+N)}\\
\cmidrule(lr){3-5}\cmidrule(lr){6-8}
Foundation & Method &
  Precision & Recall & F$_1$ &
  Precision & Recall & F$_1$ & \\
\midrule
\multirow{4}{*}{Mixtral}
& ZSCOT  & 0.831 & 0.765 & 0.792 & 0.843 & 0.822 & 0.832 & 0.812\\
& RAG    & 0.771 & 0.730 & 0.743 & 0.803 & 0.799 & 0.797 & 0.770\\
& \textbf{KEwLTM}
          & \textbf{0.857$\pm$0.022} & \textbf{0.799$\pm$0.020} & \textbf{0.822$\pm$0.010}
          & \textbf{0.857$\pm$0.011} & \textbf{0.841$\pm$0.011} & \textbf{0.847$\pm$0.008}
          & \textbf{0.835}\\
& KEwRAG & 0.792 & 0.728 & 0.746 & 0.807 & 0.814 & 0.810 & 0.778\\
\midrule
\multirow{4}{*}{Med42}
& ZSCOT  & 0.746 & 0.678 & 0.703 & 0.748 & 0.723 & 0.724 & 0.714\\
& RAG    & 0.786 & 0.748 & 0.764 & 0.760 & 0.799 & 0.759 & 0.762\\
& \textbf{KEwLTM}
          & 0.792$\pm$0.032 & 0.747$\pm$0.043 & 0.764$\pm$0.034
          & 0.785$\pm$0.021 & 0.803$\pm$0.025 & 0.787$\pm$0.026
          & 0.776\\
& KEwRAG & \textbf{0.838} & \textbf{0.793} & \textbf{0.812}
          & \textbf{0.845} & \textbf{0.849} & \textbf{0.846}
          & \textbf{0.829}\\
\bottomrule
\end{tabular}}
\end{table*}

    \subsection{Performance Comparison}\label{subsec4-1}
    Table~\ref{tab:performance-comparison} compares the performance of each method across T and N categories in terms of precision, recall, and F1 score. Note that KEwLTM’s results are averaged over eight distinct test splits (each covering 700 reports and inducing a unique LTM), whereas the other methods are evaluated once on all 800 reports.
    
    A key observation is that the relative performance of KEwRAG and KEwLTM depends on whether RAG outperforms ZSCOT for a given base model. When the base LLM does better under RAG than ZSCOT, KEwRAG also outperforms KEwLTM. Conversely, if ZSCOT yields stronger results than RAG, KEwLTM surpasses KEwRAG. This suggests that a model performing well in ZSCOT mode benefits more from KEwLTM, since robust zero-shot reasoning can produce more accurate rules in long-term memory.

    %% <MODIFICATION START>
    Although KEwLTM does not always achieve top performance, it has a notable advantage: its label-free induction process means it does not require external knowledge sources or ground-truth labels when inducing its memory. This is particularly relevant in clinical settings where such resources may be limited or access to them restricted. By contrast, standard RAG (and by extension, KEwRAG's initial retrieval step) depends on the availability and quality of external knowledge. Hence, in cases where zero-shot reasoning is already effective, KEwLTM provides a more streamlined approach, delivering a comparable (and sometimes superior) performance while avoiding the overhead and dependencies of retrieval-based methods.
    %% <MODIFICATION END>

    \subsection{Error Analysis}\label{subsec4-2} 

    To gain insight into how KEwLTM and KEwRAG benefit from induced long-term memory (or elicited rules), and to understand the types of errors made (identifying situations where these methods are weak), we performed a detailed error analysis. It is important to note that while LLMs generate textual reasoning, this reasoning can be a post-hoc rationalization and may not fully reflect the model's internal decision-making process. However, analyzing this output is valuable for identifying patterns and potential weaknesses.
    
    First, we quantified the overall error rates by determining the number of incorrectly predicted reports for all evaluated methods. Table~\ref{tab:overall_error_counts_percentages} presents these counts and the corresponding error percentages for ZSCOT, RAG, KEwLTM, and KEwRAG, applied to both Mixtral and Med42 models, across T and N stage predictions. For ZSCOT, RAG, and KEwRAG, the number of errors and percentages are based on the total 800 reports. For KEwLTM, these figures represent the average number of incorrectly predicted reports from the 700 reports in each of the eight test splits. This table provides a quantitative comparison of error propensity, defined as the proportion of incorrectly predicted reports.
    
    \begin{table*}[ht]
    \centering
    \caption{Number of incorrectly predicted reports and error percentages for T and N stage classification. Percentages are calculated as (Number of Incorrectly Predicted Reports / Total Reports Evaluated) $\times$ 100. For KEwLTM, `Num. Errors` is the average count of incorrectly predicted reports over 8 splits (700 reports per test split); for other methods, it is the count of incorrectly predicted reports out of 800.}
    \label{tab:overall_error_counts_percentages}
    \resizebox{\textwidth}{!}{%
    \begin{tabular}{@{}llcccccccc@{}}
    \toprule
    & & \multicolumn{4}{c}{Mixtral} & \multicolumn{4}{c}{Med42} \\
    \cmidrule(lr){3-6} \cmidrule(lr){7-10}
    & & \multicolumn{2}{c}{T Category} & \multicolumn{2}{c}{N Category} & \multicolumn{2}{c}{T Category} & \multicolumn{2}{c}{N Category} \\
    \cmidrule(lr){3-4} \cmidrule(lr){5-6} \cmidrule(lr){7-8} \cmidrule(lr){9-10}
    Foundation & Method & Num. Errors & Error \% & Num. Errors & Error \% & Num. Errors & Error \% & Num. Errors & Error \% \\
    \midrule
    \multirow{4}{*}{Mixtral}
    & ZSCOT  & 110 & 13.8\% & 102 & 12.8\% & 185 & 23.1\% & 210 & 26.3\% \\
    & RAG    & 148 & 18.5\% & 132 & 16.5\% & 131 & 16.4\% & 168 & 21.0\% \\
    & KEwLTM & 85.50 & 12.2\% & 82.12 & 11.7\% & 115.50 & 16.5\% & 133.50 & 19.1\% \\
    & KEwRAG & 122 & 15.3\% & 113 & 14.1\% & 97  & 12.1\% & 96  & 12.0\% \\
    \bottomrule
    \end{tabular}%
    }
    \end{table*}
    
    For a more qualitative understanding, we manually reviewed errors from the Mixtral T-stage predictions. We focused on two comparative pairs: (1) ZSCOT (baseline) versus KEwLTM (using its first test split), and (2) RAG (baseline) versus KEwRAG. In each pair, we analyzed the errors that were unique to one method when its counterpart was correct. This approach helps to highlight the specific weaknesses or differing error patterns introduced by KEwLTM and KEwRAG relative to their respective baselines. The primary error categories used for this analysis are:
    
    \begin{enumerate}
        \item \textbf{Incorrect Information Extraction (IIE)}: The model missed or misread key facts in the report.
        \item \textbf{Incorrect Inference (Inf.)}: The model identified the relevant facts but drew an incorrect conclusion. This is a general category for logical errors not covered by NI or IK.
        \item \textbf{Numerical Incompetence (NI)}: A specific type of incorrect inference where the model made errors in numerical comparison or simple arithmetic (e.g., reading 1.9 cm as "$>$ 2 cm").
        \item \textbf{Incorrect Knowledge (IK)}: Another specific type of incorrect inference where the model explicitly stated or used an incorrect domain rule (e.g., an incorrect AJCC rule for staging).
        \item \textbf{Conflicting Ground Truth (CGT)}: The provided ground-truth label appeared to contradict the information available in the pathology text, suggesting potential dataset issues.
        \item \textbf{Incomplete Information (IncInf.)}: The report itself lacked sufficient detail to definitively determine the cancer stage according to guidelines.
    \end{enumerate}

    \begin{table*}[ht]
    \centering
    \caption{Distribution of unique error causes in comparative pairs (Mixtral T-Stage). Each row shows errors made by one method when its counterpart in the pair was correct. Error categories: Incorrect Information Extraction (IIE), general Incorrect Inference (Inf.), Numerical Incompetence (NI), Incorrect Knowledge (IK), Conflicting Ground Truth (CGT), Incomplete Information (IncInf.).}
    \label{tab:qualitative_unique_error_distribution}
    \resizebox{\textwidth}{!}{%
    \begin{tabular}{@{}lccccccc|c@{}}
    \toprule
    Error Set (Method-Only Errors when Baseline was Correct) & IIE & Inf. & NI & IK & CGT & IncInf. & Total Unique Errors \\
    \midrule
    ZSCOT-only errors (vs. KEwLTM)    & 10 & 6  & 24 & 5  & 1 & 0 & 46 \\
    KEwLTM-only errors (vs. ZSCOT)   & 4  & 0  & 19 & 2  & 1 & 0 & 26 \\
    \midrule
    RAG-only errors (vs. KEwRAG)      & 11 & 5  & 53 & 11 & 1 & 0 & 81 \\
    KEwRAG-only errors (vs. RAG)     & 18 & 5  & 19 & 10 & 2 & 1 & 55 \\
    \bottomrule
    \end{tabular}%
    }
    \end{table*}

    Table~\ref{tab:qualitative_unique_error_distribution} presents the distribution of these error causes for the unique errors identified in our comparative pairs. For the 'Incorrect Knowledge (IK)' category, a key observation is that while the induced LTM in KEwLTM, the elicited rules in KEwRAG, or the retrieved chunks in standard RAG generally contained correct domain knowledge, errors could still arise. 
    \begin{itemize}
        \item \textit{Example of Incorrect Knowledge in KEwLTM(-only error):}
        \begin{itemize}
            % \item \textbf{Patient ID:} TCGA-AO-A0JA.2E8A1816-E963-4063-B2F0-6289514084A5
            \item \textbf{Pathology Report Excerpt:} "[...] HISTOLOGIC GRADE II/III (MODERATE TUBULE FORMATION), NUCLEAR GRADE II/III. (MODERATE VARIATION IN SIZE AND SHAPE). MEASURING `3.6 CM IN LARGEST. [...] TUMOR \#2: - INVASIVE DUCTAL CARCINOMA, NOS TYPE, WITH FOCAL LOBULAR GROWTH. PATTERN, HISTOLOGIC GRADE II/III (MODERATE TUBULE FORMATION) , NUCLEAR GRADE. II/III (MODERATE VARIATION IN SIZE AND SHAPE) . MEASURING 2.5 CM IN LARGEST. [...]"
            \item \textbf{Ground Truth:} T2
            \item \textbf{KEwLTM's Reasoning:} "The report indicates that there are two tumors in the right breast, with the largest one measuring 3.6 cm and the second one measuring 2.5 cm. This meets the criteria for T3 stage as the tumor size is more than 5 cm (3.6 cm + 2.5 cm = 6.1 cm). Additionally, the report mentions that there is extensive vascular invasion and extranodal extension ($>$2mm) of carcinoma. These factors also support the T3 stage."
            \item \textbf{Analysis:} The model correctly extracted tumor sizes but erroneously applied a non-existent AJCC rule by summing the diameters of multifocal tumors to determine the T-stage. This demonstrates an application of incorrect domain knowledge by KEwLTM.
        \end{itemize}
        \vspace{0.5em}
        \item \textit{Example of Incorrect Knowledge in KEwRAG(-only error):}
        \begin{itemize}
            % \item \textbf{Patient ID:} TCGA-EW-A424.EE644ADF-639E-450A-92E6-2A6F1BAD561F
            \item \textbf{Pathology Report Excerpt:} "F. Pathologic Interpretation: A. RIGHT BREAST TOTAL MASTECTOMY: INVASIVE LOBULAR CARCINOMA, moderately differentiated, Nottingham grade 2 (3+2+2=7), 5.5 cm in. greatest dimension. Margins are negative for carcinoma, closest margin is posterior (3.5 cm). Negative for lymphovascular invasion. Lobular carcinoma in situ is present. Skin and nipple with no significant pathologic changes. […]"
            \item \textbf{Ground Truth:} T3
            \item \textbf{KEwRAG's Reasoning:} "According to the report, the patient had an invasive lobular carcinoma that was 5.5 cm in greatest dimension. [...] T3: This stage includes invasive breast carcinomas that are more than 5 cm in size. It also includes tumors of any size with direct extension to the chest wall or skin. The report states that the tumor is 5.5 cm in size and does not mention any direct extension to the chest wall or skin. Therefore, it does not meet the criteria for T3. [...] Therefore, based on the rules provided and the information in the report, the patient's pathologic T stage is T2."
            \item \textbf{Analysis:} KEwRAG correctly identified the tumor size (5.5 cm) but misstated the AJCC T3 criteria, incorrectly claiming that T3 requires \textit{both} size $>$ 5 cm \textit{and} direct extension (whereas size $>$ 5 cm alone is sufficient for T3). This explicit misstatement of the rule led to an incorrect downstaging to T2.
        \end{itemize}
    \end{itemize}
    \vspace{0.5em}

    'Numerical Incompetence (NI)' was a prevalent issue, across different methods, highlighting a general challenge for LLMs in precise numerical comparisons, even when the underlying rules were understood. 
    
    \begin{itemize}
        \item \textit{Example of Numerical Incompetence in ZSCOT(-only error):}
        \begin{itemize}
            % \item \textbf{Patient ID:} TCGA-AR-A2LK.73FFBADB-37C0-4F17-8C95-F8DC94EF1AD3
            \item \textbf{Pathology Report Excerpt:} "Final Diagnosis. Breast, right, wide local excision: Infiltrating lobular carcinoma, Nottingham grade II (of. III)[tubules 3/3, nuclei 2/3, mitoses 1/3; Nottingham score 6/9], forming a 5.2 x 3.0 x 2.6 cm. mass. Lobular carcinoma in situ is present and represents less than 5\% of the overall tumor. volume. Angiolymphatic invasion is present. The non-neoplastic breast parenchyma show. nonproliferative fibrocystic changes. [...]"
            \item \textbf{Ground Truth:} T3
            \item \textbf{Model's Reasoning (ZSCOT):} "The report states that the largest contiguous invasive carcinoma measures 5.2 x 3.0 x 2.6 cm, which puts it in the T2 category (more than 2 cm but not more than 5 cm in size). Additionally, there is no mention of invasion into the skin, chest wall, or underlying muscles, which would have upstaged the tumor to a T3 or T4."
            \item \textbf{Analysis:} The model correctly extracted the tumor size (5.2 cm) and cited the proper AJCC rule boundary (T2 vs. T3 at 5 cm) but still incorrectly classified 5.2 cm as "not more than 5 cm," thus misapplying a simple numeric comparison.
        \end{itemize}
    \end{itemize}
    \vspace{0.5em}
    
    'Incorrect Information Extraction (IIE)' also occurred, sometimes manifesting as the LLM overlooking explicitly stated information in the report, even when such information was crucial for correct staging.
    \begin{itemize}
        \item \textit{Example of Incorrect Information Extraction in KEwRAG(-only error):}
        \begin{itemize}
            % \item \textbf{Patient ID:} TCGA-D8-A1XC.E5D2E429-C41C-47E2-96CB-F3B6F412499B
            \item \textbf{Pathology Report Excerpt:} "[...] page 2 / 2. cxamination No.: Gender: F. Complementary diagnosis: Complementary macroscopic description: Extensive skin proliferation. Minor ulceration in the skin. Skin proliferation confirmed microscopically. Focal. tumour pattern along the incision line at the side of the base and lower boundary. Test result: Carcinoma mucinosum invasivum with a neuroendocrinic component mammae dextrae. Micrometastasis. carcinomatosa in lymphonodo axillae (No 1/XX) (NHG2, pT4b, pNIa mi). [...]"
            \item \textbf{Ground Truth:} T4
            \item \textbf{KEwRAG's Reasoning:} "The report mentions the size of the tumor as 7 x 4 x 8 cm. According to rule 3, T3 stage includes invasive breast carcinomas that are more than 5 cm in size. The largest dimension of the tumor is 8 cm, which is more than 5 cm, therefore the T stage is T3."
            \item \textbf{Analysis:} The report explicitly stated "pT4b" and described "Extensive skin proliferation. Minor ulceration in the skin," which are key features for T4b. However, the model focused only on the tumor size to infer T3 and seemingly ignored or failed to extract these critical pieces of information.
        \end{itemize}
    \end{itemize}
    
    These examples illustrate that errors can arise not only from failing to retrieve or elicit correct knowledge but also from misapplying correct knowledge, struggling with numerical comparisons, or failing to extract salient information from the text. 
    
    Across both comparisons, Numerical Incompetence stands out as a frequent cause of unique errors for all methods, emphasizing the challenge LLMs face with precise numerical tasks. Incorrect Information Extraction is also a common issue. For example, some IIE errors arose when pathology reports contained explicit mentions of the cancer stage that the LLM seemingly failed to prioritize or directly utilize, instead attempting to infer the stage from other textual details. It's a common behavior pattern for LLMs to sometimes overlook direct information if the surrounding context is complex or if their prompting/training steers them towards more inferential paths.
    
    In summary, while both KEwLTM and KEwRAG demonstrate strong performance, the error analysis reveals persistent challenges, particularly in numerical reasoning. Future work will explore strategies to specifically address these error categories, such as integrating external calculation tools.

\section{Discussion}\label{sec5}
    \subsection{Implication}\label{subsec5-1}
    %% <MODIFICATION START>
    Based on the experimental results and analysis, KEwLTM and KEwRAG offer complementary strengths in automating cancer staging, moving beyond straightforward applications of existing prompting techniques by incorporating specific mechanisms for rule induction and application tailored for clinical needs.

    On one hand, KEwLTM capitalizes on the model’s zero-shot reasoning ability. Its distinct contribution lies in inducing interpretable, domain-specific rules directly from a small set of unannotated training reports—critically, without requiring ground-truth labels or external data. This label-free induction process makes KEwLTM particularly suitable for clinical settings where annotated data is scarce or privacy-restricted. It enhances transparency and cost-effectiveness, and reduces the overhead typical of retrieval-based approaches. For LLMs already adept at ZSCOT, KEwLTM’s rule-induction can lead to higher recall and balanced performance.

    By contrast, KEwRAG is better suited to scenarios where the base LLM benefits from external retrieval. Unlike typical RAG workflows, which often feed raw text “chunks” into the LLM for each query, KEwRAG’s novelty is in distilling these chunks into explicit, structured rules up front. This one-time rule synthesis turns scattered text into a coherent, rule-based context that can be applied repeatedly without further retrieval calls. This approach offers clearer auditing and traceability: clinicians and developers can review precisely which rules were extracted from the guidelines and how they guide the model’s inference, leading to a more transparent and interpretable workflow than appending raw text snippets. This also reduces computational overhead associated with per-query retrieval.

    Rule generation in KEwLTM and KEwRAG rests on two different philosophies. KEwLTM's iterative refinement from unannotated reports reflects a learning process. In contrast, KEwRAG performs a one-time knowledge distillation from an authoritative external source. Both approaches, however, yield an explicit, interpretable rule set, a critical asset in clinical settings where explainability is paramount. In practice, the choice between KEwLTM and KEwRAG (or a hybrid) will likely hinge on:
    %% <MODIFICATION END>
    
    \begin{enumerate}
        \item \textbf{Base-Model Competency}: Models already strong in zero-shot reasoning can exploit KEwLTM more effectively, while weaker zero-shot models may need KEwRAG’s external knowledge.
        \item \textbf{Resource Constraints}: KEwLTM’s lower computational overhead and independence from retrieval systems make it appealing for resource-limited environments.
        \item \textbf{Transparency vs. Performance Trade-offs}: KEwRAG’s more complex retrieval pipeline may occasionally offer higher accuracy, but can introduce additional complexity in terms of maintenance and optimization.
    \end{enumerate}

    %% <CHANGE START 0416>
    It is important to contextualize the performance observed here, particularly for the baseline methods, within the broader scope of the TCGA dataset. Our focus on the BRCA subset allowed for a detailed exploration of the novel KEwLTM and KEwRAG methods. However, related work by Chang et al.~\cite{Chang2024-qd} evaluated the Med42-70B model using the ZSCOT approach (equivalent to our ZSCOT baseline) on the standard 15\% test split of the full TCGA dataset, encompassing multiple cancer types. Their findings provide valuable benchmarks: Med42-70B with ZSCOT achieved competitive performance against a fine-tuned Clinical-BigBird model on the full test set, achieving a macro F1 of 0.82 for the N category and 0.78 for the T category. Specifically for the BRCA subset within their test data, Chang et al.~\cite{Chang2024-qd} reported macro F1 scores of 0.72 for T and 0.80 for N using Med42+ZSCOT. These figures align reasonably well with our ZSCOT baseline results for Med42 on BRCA (Table~\ref{tab:performance-comparison}: T Macro F1=0.703, N Macro F1=0.724), confirming the general capability of ZSCOT on this task and providing a reference point against which the improvements offered by KEwLTM and KEwRAG in specific scenarios can be assessed.
    %% <CHANGE END 0416>
    %% <MODIFICATION START>
    
    Ultimately, both approaches highlight how LLMs can be guided to learn and apply domain-specific rules using minimal data. This can be done without relying on extensive annotated datasets or complex fine-tuning. Our methods are adaptations of iterative prompting and Retrieval-Augmented Generation (RAG). Specifically, KEwLTM uses a label-free iterative process to induce rules. KEwRAG, on the other hand, employs single-shot retrieval followed by the extraction of interpretable rules. Both approaches explicitly focus on deployment in clinical settings where data is minimal. These particular adaptations represent meaningful contributions. They help connect LLMs with real-world medical tasks, paving the way for more scalable and interpretable AI solutions in healthcare.
    %% <MODIFICATION END>
    
%% Revision start <0514>
\subsection{Limitations}\label{subsec5-2} 
The findings presented in this study, while promising for the specific context investigated, should be considered in light of several limitations. Firstly, our experimental evaluation was conducted exclusively on breast cancer (BRCA) pathology reports sourced from The Cancer Genome Atlas (TCGA) project. This focus on a single cancer type and a single data repository inherently restricts the direct generalizability of our findings to other types of cancer or to clinical environments and datasets beyond TCGA. While we chose BRCA due to its prevalence and data availability within TCGA, we acknowledge that this does not substantiate performance on other malignancies.

Secondly, variations in pathology reporting styles, specific terminology, and the application or version of staging guidelines (e.g., different editions of the AJCC manual) can be significant across different cancer types and particularly across different medical institutions. Our current study did not encompass such heterogeneity. Consequently, the performance of KEwLTM and KEwRAG could be impacted when applied to reports from diverse institutional settings or for other cancer types, which may present distinct linguistic characteristics, rule complexities, or formatting. For instance, the rule elicitation process in KEwLTM might require adaptation or a more diverse set of initial examples if applied to a new domain with substantially different reporting norms, and the effectiveness of KEwRAG would depend on the availability and nature of external guidelines for other cancer types.

Therefore, while our methods show promise for automated staging on TCGA BRCA reports, their robustness and effectiveness across a broader range of oncological data remain to be rigorously validated. As outlined in our future work (Section~\ref{subsec5-3}), crucial next steps involve validating and potentially adapting these approaches on more diverse datasets. This includes testing on reports for other cancer types and on data sourced from multiple different institutions to assess real-world applicability and address the challenge of inter-institutional variability.
%% Revision end <0514>
    \subsection{Future Direction}\label{subsec5-3}
        \subsubsection{Improving Memory Accuracy and Reducing Hallucination}\label{subsubsec5-3-1}
        A key area for future work is to enhance the accuracy of KEwLTM’s induced long-term memory while minimizing hallucinations. We plan to incorporate reinforcement-based feedback mechanisms into the prompting process, encouraging the LLM to critique and refine its own outputs~\cite{Madaan2023-om}. Such self-reflection strategies may help reduce erroneous or fabricated content and further solidify the model’s grasp of domain-specific rules.

        \subsubsection{Addressing Numerical Incompetence and External Tool Integration}\label{subsubsec5-3-2}
        Error analyses revealed that the LLM’s current difficulty in handling numerical tasks, or "numerical incompetence" as discussed in Section~\ref{subsec4-2} and highlighted by Mahendra et al.~\cite{mahendra-etal-2024-numbers}, occasionally undermines performance. This is particularly true in cases requiring precise comparisons or calculations, which are critical for tasks like dosage calculation or nuanced data interpretation in clinical settings. Although human oversight can resolve these issues, there is clear room for improvement in the model's autonomous capabilities.

        One promising way to mitigate such numerical reasoning errors is through tool calling or function calling. In this approach, the LLM delegates arithmetic or other precise computations to an external, specialized function. For example, instead of performing all logic internally, the LLM’s role is to parse out the numbers and determine what kind of operation (e.g., comparison, threshold check) is needed. The model then generates a structured call to an external function (e.g., a Python function) with these extracted values. The external function is designed to map numerical values to the correct T or N category deterministically, according to predefined thresholds in the AJCC guidelines. This offloads the potential for arithmetic mistakes from the LLM to a concise, fully testable function. Once the external tool returns a result (e.g., `T1`), the LLM can seamlessly incorporate that outcome into its final response or reasoning chain. Since the numeric logic is handled by code, the risk of the model “hallucinating” or miscalculating a threshold is greatly reduced.
        
        By splitting linguistic reasoning (handled by the LLM) from precise numeric operations (handled by a specialized module), we can gain the benefits of both: the flexibility and language understanding of an LLM, and the reliability of deterministic arithmetic. This approach is increasingly adopted in various “LLM + Tools” frameworks (e.g., LangChain, OpenAI function calling) and is particularly relevant to clinical text processing, where accurate numeric calculations are critical for tasks like stage classification and drug dosage calculations.

        \subsubsection{Expanding to Other Cancer Types and Clinical Tasks}\label{subsubsec5-3-3}

        A critical next step, essential for validating the broader applicability and generalizability of our proposed methods beyond the initial proof-of-concept presented here, involves evaluating both KEwLTM and KEwRAG across a diverse range of cancer types within the TCGA dataset and potentially other clinical datasets. This expansion is necessary to confirm whether the promising results observed in the BRCA cohort translate to reports with different structures, terminologies, and staging criteria inherent to other malignancies. Furthermore, to effectively manage the heterogeneity across different cancer types, we plan to investigate the development of specialized LLM agents. Each agent could potentially leverage KEwLTM or KEwRAG to induce and maintain a long-term memory or rule set specific to a particular cancer type. Examining how these specialized agents might collaborate~\cite{Liang2023-km}, sharing insights or delegating tasks, could offer a scalable and robust approach to accurately determining cancer stages across the full spectrum of pathology reports encountered in real-world clinical settings.
        %% <CHANGE END 0416>

        % Revision start (0512)
        \subsubsection{Optimizing the KEwRAG Pipeline}\label{subsubsec5-3-4}
        Further investigation into optimizing the KEwRAG pipeline is warranted. This includes systematically evaluating the impact of the number of retrieved chunks ($k$) and experimenting with more sophisticated query formulations (e.g., targeted queries for specific staging components like T, N, or substages) to potentially improve the quality of the retrieved context. Furthermore, while our current KEwRAG design uses single-pass synthesis for efficiency, exploring an iterative refinement mechanism for KEwRAG, where the LLM could potentially refine the elicited rules based on the retrieved chunks over multiple steps, could be a valuable direction to enhance rule quality.
        % Revision end (0512)

\section{Conclusion}\label{sec6}
%% <MODIFICATION START>
In this work, we investigated two novel prompting approaches—Knowledge Elicitation with Long-Term Memory (KEwLTM) and Knowledge Elicitation with Retrieval-Augmented Generation (KEwRAG)—to automate the extraction of pathologic T and N stages from free-text pathology reports. While building upon established concepts, our methods introduce specific adaptations to enhance utility in clinical settings. KEwLTM employs a label-free iterative process to induce domain-specific rules directly from a small subset of unannotated training reports. KEwRAG modifies standard RAG by performing an upfront retrieval and synthesis of explicit rules from external guidelines, which are then consistently applied. Both methods aim to reduce reliance on large labeled datasets and iterative fine-tuning, while prioritizing the generation of interpretable rule sets.

Our experiments on breast cancer pathology reports from the TCGA dataset revealed that KEwLTM excels when the underlying large language model (LLM) demonstrates strong zero-shot reasoning, leveraging its ability to derive rules without external labels. KEwRAG delivers better outcomes for models that benefit from external retrieval, with its one-shot rule synthesis providing a transparent and auditable knowledge base while avoiding repeated lookup overheads.

Despite these strengths, our methods have limitations. KEwLTM’s performance depends on the quality of the base model’s zero-shot reasoning for effective rule induction. KEwRAG relies on the availability and correctness of external resources for its initial rule synthesis. The practical deployment of both methods requires careful integration into clinical workflows.

Looking ahead, future work should focus on validating these methods across a broader range of cancer types and diverse clinical datasets to assess generalizability. Expanding these methods to encompass more complex scenarios, exploring adaptive rule induction, and refining techniques for hybrid frameworks are also important directions. Deeper error analyses on real-world reports will further guide the integration of such LLM-driven workflows into clinical decision-support.

Overall, our results demonstrate that LLMs, guided by tailored knowledge elicitation strategies like KEwLTM and KEwRAG, can learn and apply domain-specific rules with minimal data and expert supervision. These approaches represent meaningful contributions by adapting existing techniques for enhanced interpretability and practicality in healthcare, opening a path to more scalable, transparent, and trustworthy AI-driven solutions in complex clinical tasks.
%% <MODIFICATION END>

\backmatter

\bmhead{Acknowledgements}
This work was supported in part by the National Science Foundation under the Grants IIS-1741306 and IIS-2235548, and by the Department of Defense under the Grant DoD W91XWH-05-1-023.  This material is based upon work supported by (while serving at) the National Science Foundation.  Any opinions, findings, and conclusions or recommendations expressed in this material are those of the author(s) and do not necessarily reflect the views of the National Science Foundation.

\section*{Declarations}
\subsection*{Funding}
This work was funded by the National Science Foundation under the Grants IIS-1741306 and IIS-2235548, and by the Department of Defense under the Grant DoD W91XWH-05-1-023.  
\subsection*{Competing interests}
Author Christopher C. Yang (the corresponding author) is the editor-in-Chief of this journal. The authors have no other competing interests to declare.
\subsection*{Ethics approval and consent to participate}
Not applicable
% \subsection*{Consent for publication}
% \subsection*{Data availability}
% \subsection*{Materials availability}
% \subsection*{Code availability}
% \subsection*{Author contribution}

\bibliography{sn-bibliography}% common bib file

\backmatter
\appendix
\section{Hyperparameter Sensitivity Analysis for KEwLTM}\label{app:sensitivity}

The analyses presented in this appendix, including all figures and discussions regarding the impact of the number of training reports and edit distance thresholds, were conducted for the KEwLTM method using the Mixtral-8x7B-Instruct-v0.1 model.

\subsection{Impact of Number of Training Reports on Performance}
\begin{figure}[h!]
    \centering
    \includegraphics[width=0.8\textwidth]{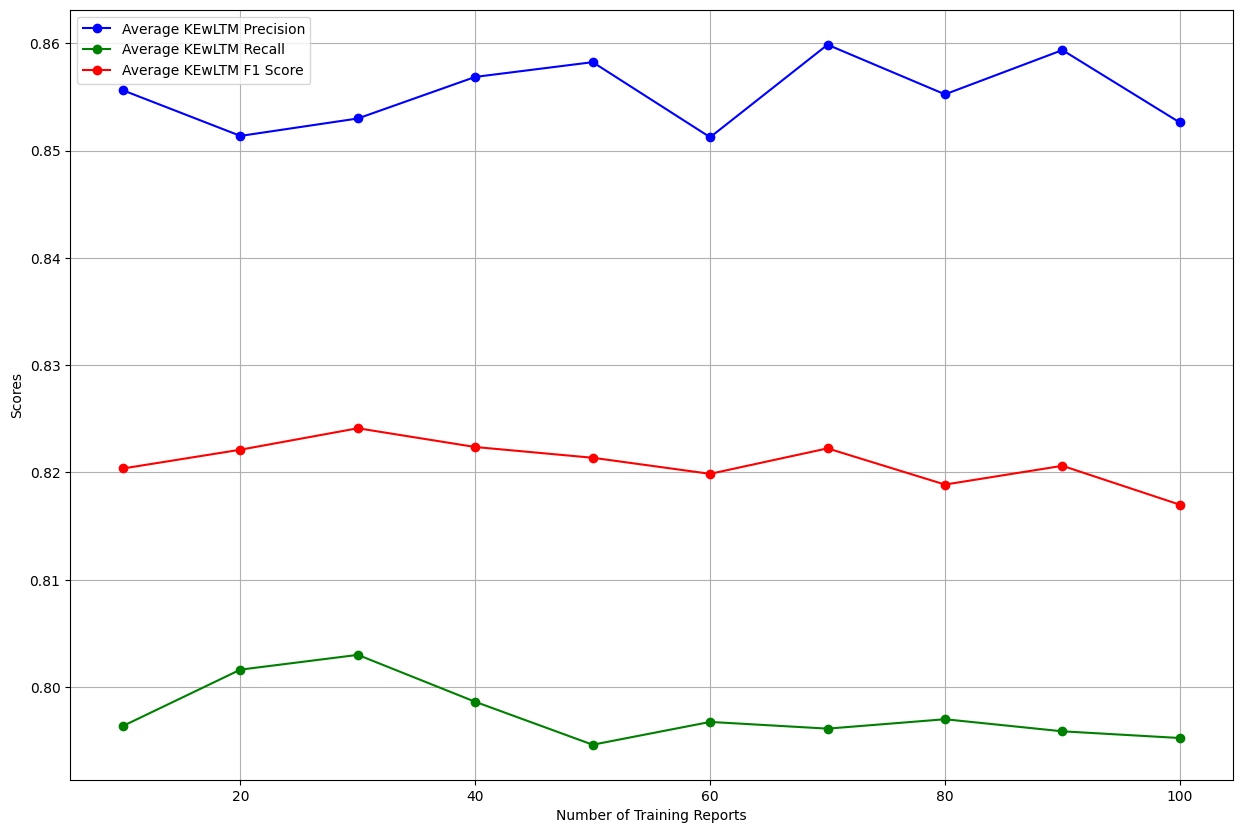} 
    \caption{Impact of the number of training reports on KEwLTM performance for T-category classification. The plots show average Precision, Recall, and F1-score over eight random test splits.}
    \label{fig:num_training_t}
\end{figure}

\begin{figure}[h!]
    \centering
    \includegraphics[width=0.8\textwidth]{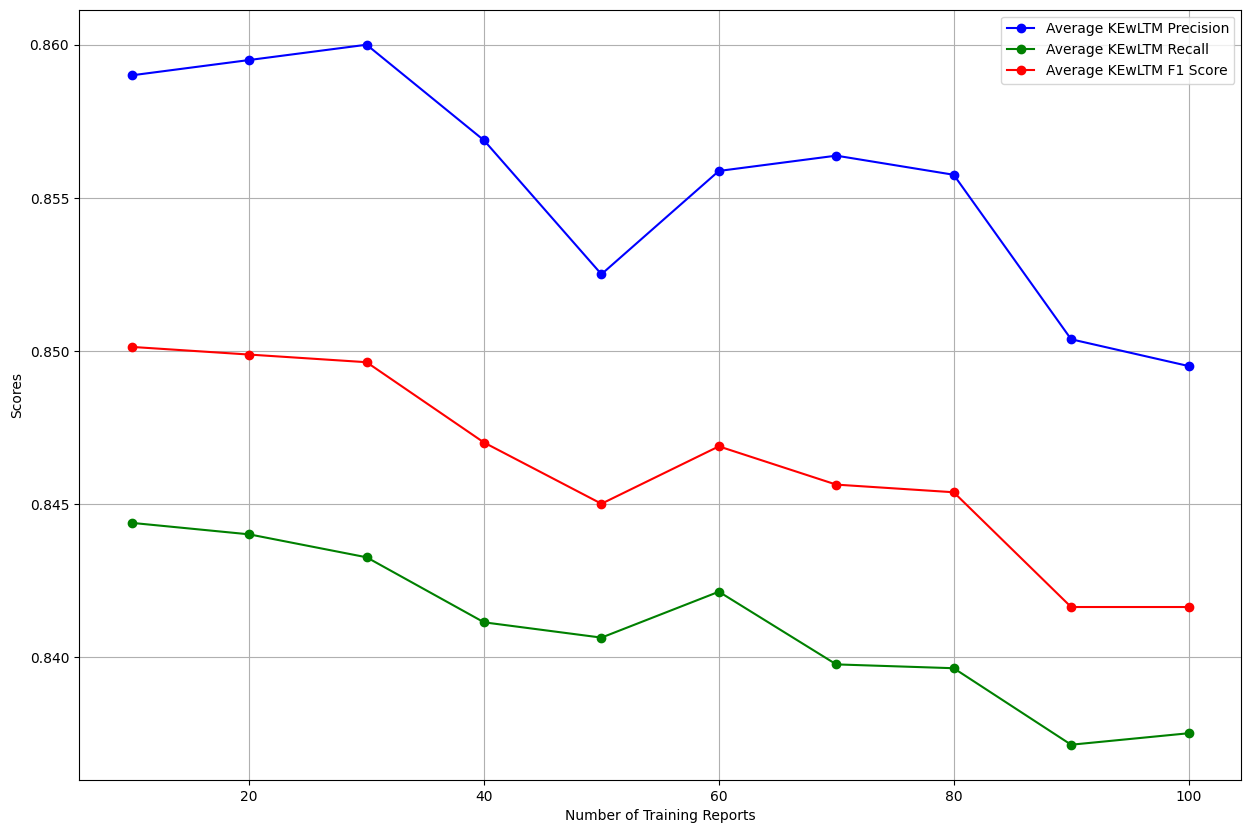} 
    \caption{Impact of the number of training reports on KEwLTM performance for N-category classification. The plots show average Precision, Recall, and F1-score over eight random test splits.}
    \label{fig:num_training_n}
\end{figure}

In this section, we analyze the impact of varying the number of training reports used for inducing the long-term memory in KEwLTM on its final classification performance. Figure~\ref{fig:num_training_t} and Figure~\ref{fig:num_training_n} illustrate the average Precision, Recall, and F1-score for T-category and N-category classification, respectively, as the number of training reports varies from 10 to 100 in increments of 10.

As observed in Figure~\ref{fig:num_training_t} (T-category) and Figure~\ref{fig:num_training_n} (N-category), the F1-scores do not always show a single, distinct peak. Our choice of 40 training reports for the main experiments was based on selecting a point that provided a consistently robust and representative performance across both T and N categories. The number offered a fair and balanced performance level, avoiding extremes. This approach was intended to provide a reasonable basis for evaluating KEwLTM's general effectiveness.

\subsection{Impact of Edit Distance Threshold on Memory Evolution}
\begin{figure}[h!]
    \centering
    \includegraphics[width=0.8\textwidth]{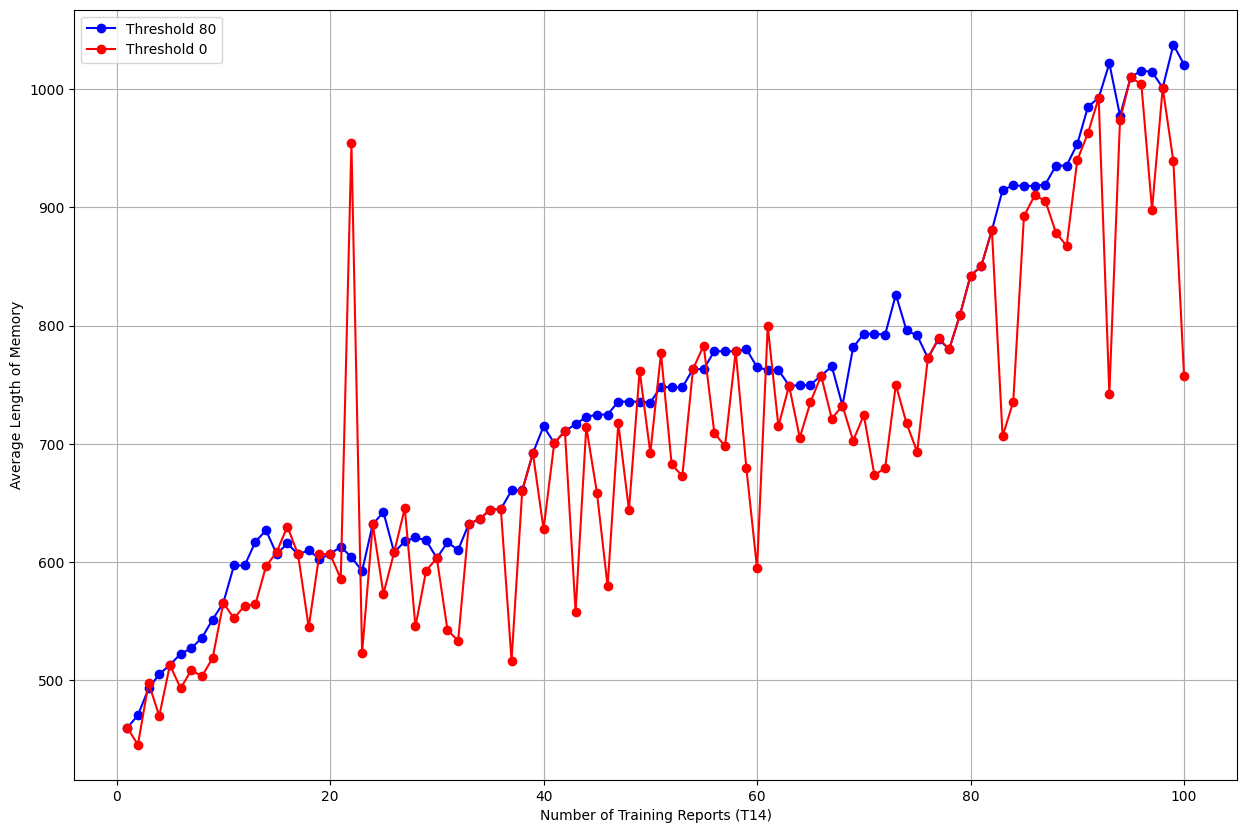} 
    \caption{Average long-term memory length evolution for T-category rules with different edit distance threshold conditions during KEwLTM memory induction.}
    \label{fig:edit_distance_t}
\end{figure}

\begin{figure}[h!]
    \centering
    \includegraphics[width=0.8\textwidth]{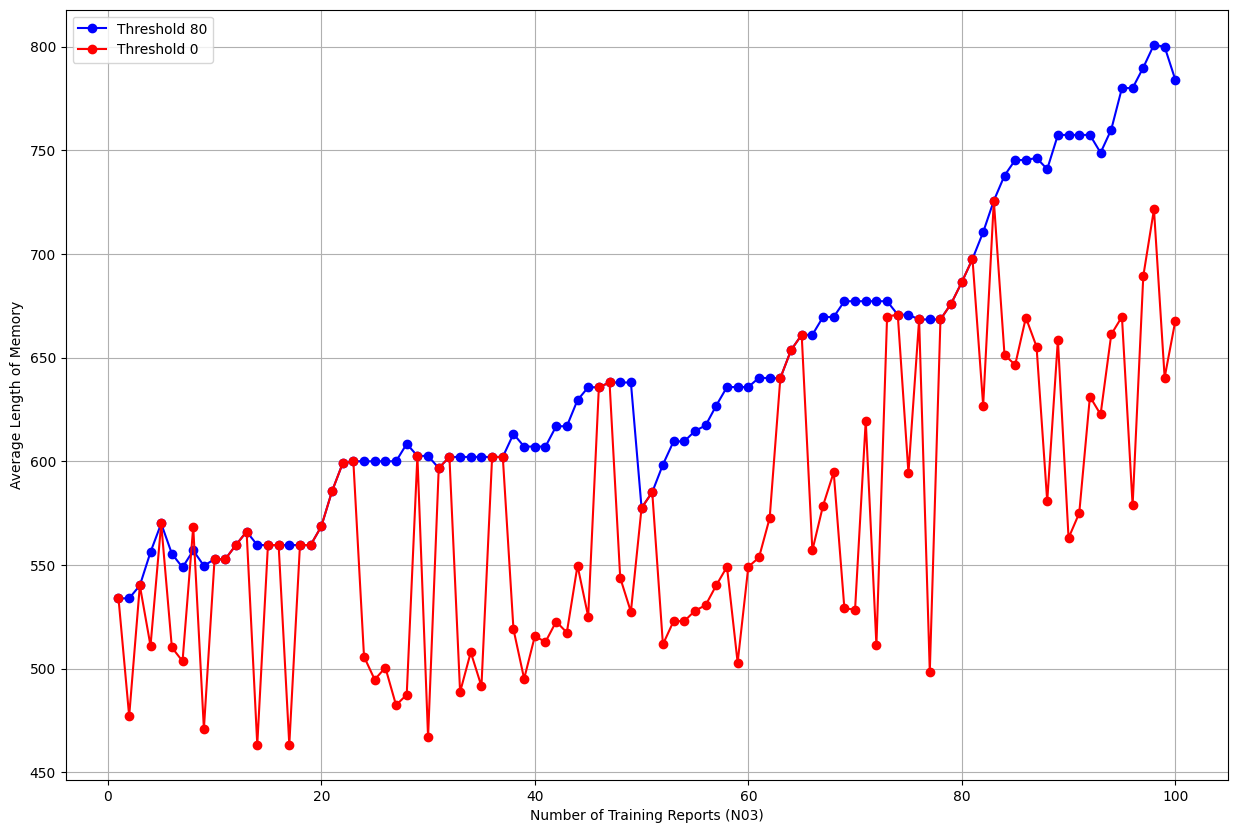}
    \caption{Average long-term memory length evolution for N-category rules with different edit distance threshold conditions during KEwLTM memory induction.}
    \label{fig:edit_distance_n}
\end{figure}

The edit distance threshold plays a role in controlling how the long-term memory $\mathcal{M}$ in KEwLTM is updated. A threshold ensures that updates to the memory are only accepted if the newly proposed memory is sufficiently similar to the existing one, promoting stable and incremental evolution. Figure~\ref{fig:edit_distance_t} and Figure~\ref{fig:edit_distance_n} show how the average length of the induced memory for T and N categories, respectively, changes as more training reports are processed. These plots compare two conditions: using a similarity threshold of 80 (meaning updates are accepted if the Levenshtein distance is $\leq 20$ compared to the previous memory) versus using a threshold of 0 (allowing any change or representing no filtering).

As illustrated, applying a similarity threshold of 80 generally leads to a more gradual and stable growth in long-term memory length compared to using no effective threshold (Threshold 0), where the long-term memory length can exhibit more pronounced fluctuations. This stabilization is important for developing a consistent set of rules. Our choice of a threshold of 80 for the main experiments (as discussed in Section~\ref{subsubsec3-4-3}) was based on this observation that it encourages a more controlled memory update process, leading to the stabilization of the long-term memory content after exposure to a certain number of reports (around 40, in conjunction with the number of reports analysis).
\end{document}